\pdfoutput=1

\documentclass[11pt]{article}

\ifx\review\undefined
\usepackage[]{acl}
\else
\usepackage[review]{acl}
\fi

\usepackage{amsmath}
\usepackage{amssymb}
\usepackage{geometry}
\usepackage{times}
\usepackage{latexsym}
\usepackage[T1]{fontenc}
\usepackage[utf8]{inputenc}
\usepackage{microtype}
\usepackage{inconsolata}
\usepackage{graphicx}
\usepackage[]{todonotes}
\usepackage{booktabs}
\usepackage{makecell}
\usepackage{enumitem}
\usepackage{url}
\usepackage{fontawesome}
\usepackage{pifont}
\usepackage{array}
\usepackage{multirow}
\usepackage{xcolor}
\usepackage{colortbl}
\usepackage{subcaption}
\usepackage[belowskip=-0.2em,labelfont=bf,labelsep=endash]{caption}
\usepackage[round-mode=places,round-precision=0,group-separator={,},output-decimal-marker={.}]{siunitx}
\usepackage{tikz}
\usetikzlibrary{calc}
\usepackage{pgfplots}
\pgfplotsset{compat=1.18}

\graphicspath{{figures/}}

\newcommand{\lagom}{L\textsuperscript{\hspace{-3pt}A}G\textsubscript{O}M$\boldsymbol{\cdot}$NLP}

\newcommand{\eg}[0]{\textit{e.g.,}}

\newcommand{\model}{\mathcal{M}}
\newcommand{\mmono}{\model^{\text{Mono}}}
\newcommand{\mmulti}{\model^{\text{Multi}}}

\newcommand{\langs}{\mathcal{L}}
\newcommand{\lang}{L}

\newcommand{\vocab}{V}

\newcommand{\corpus}{\mathcal{C}}
\newcommand{\cmono}{\corpus^{\text{Mono}}}
\newcommand{\cmonotest}{\corpus^{\text{Mono'}}}
\newcommand{\cmulti}{\corpus^{\text{Multi}}}

\newcommand{\calttest}{\corpus^{\text{Alt'}}}

\newcommand{\tokenizer}{\mathcal{T}}
\newcommand{\tmulti}{\tokenizer^{\text{Multi}}}
\newcommand{\tmono}{\tokenizer^{\text{Mono}}}

\newcommand{\nllarrow}{\text{NLL} \downarrow}
\newcommand{\mrrarrow}{\text{MRR} \uparrow}
\newcommand{\bpcarrow}{\text{BPC} \downarrow}

\newcommand{\de}{\text{DE}}
\newcommand{\en}{\text{EN}}

\newcommand{\inconsistent}{{\textcolor{inconsistent}{$\times$}}}
\newcommand{\consistent}{{\textcolor{consistent}{\checkmark}}}

\definecolor{consistent}{HTML}{589D54}
\definecolor{inconsistent}{HTML}{9F1010}
\definecolor{over}{HTML}{4C76B3}
\definecolor{under}{HTML}{F68E23}

\title{Form and Meaning in Intrinsic Multilingual Evaluations}
\author{Wessel Poelman \and Miryam de Lhoneux\\
    \lagom, Department of Computer Science, KU Leuven\\ 
    \texttt{\{firstname.lastname\}@kuleuven.be}
}

\begin{document}
\maketitle

\begin{abstract}

Intrinsic evaluation metrics for conditional language models, such as perplexity or bits-per-character, are widely used in both mono- and multilingual settings.
These metrics are rather straightforward to use and compare in monolingual setups, but rest on a number of assumptions in multilingual setups.
One such assumption is that comparing the perplexity of CLMs on parallel sentences is indicative of their quality since the \emph{information content} (here understood as \emph{the semantic meaning}) is the same.
However, the metrics are inherently measuring \emph{information content} in the \emph{information-theoretic} sense.
Consistency in meaning does not neutralize different forms (paraphrases).
We make such assumptions explicit and discuss their implications.
We perform experiments with six metrics on two multi-parallel corpora both with mono- and multilingual models.
We find that current metrics are not universally comparable and look at the form-meaning debate to provide some explanation for this.

\end{abstract}

\section{Introduction}\label{sec:intro}
Intrinsic evaluation metrics, such as perplexity \cite[PPL; ][]{jelinek1977perplexity}, are often the first step in evaluating conditional language models (CLMs).
PPL is the most ubiquitous, but other transformations of the loss (or negative log-likelihood) are also used.
Intrinsic evaluations are sometimes the only option for certain languages due to resource availability \cite{joshi2020state}.

There are two multilingual setups where these metrics are used: \emph{a single multilingual model} or \emph{multiple monolingual models}.
In the multilingual model setting the intrinsic metrics indicate how well a single model learns multiple languages.
Comparing metrics is used for model design: \emph{model A is better than B since A achieves a lower PPL}. 
For the setting with multiple monolingual models, the metrics are used to study the interaction of language characteristics and language modeling: \emph{which language is hard to model?}
The comparisons are used to describe languages:
\emph{the PPL for language A is lower than B, A is therefore easier to model}.

\begin{figure}
    \centering
    \resizebox{\linewidth}{!}{\begin{tikzpicture}[
    baseline/.style={thin, gray!80},
    de_point/.style={circle, fill=over, inner sep=2.1pt, opacity=0.7},
    en_point/.style={circle, draw=black, fill=under, inner sep=2.8pt, thick},
    label_style/.style={font=\normalsize},
    check/.style={color=consistent, font=\Large},
    cross/.style={color=inconsistent, font=\Large}
]
    \begin{scope}[yshift=3cm, xscale=1.2]
        \node[left, label_style] at (-0.5, 0) {NLL};
        \draw[baseline] (0,0) -- (7,0);
        \foreach \x in {4.45, 4.63, 6.30, 6.36} \node[de_point] at (\x, 0) {};
        \node[en_point] at (2.19, 0) {};
        \node[check, anchor=west] at (7.2, 0) {\checkmark};
    \end{scope}

    \begin{scope}[yshift=2cm, xscale=1.2]
        \node[left, label_style] at (-0.5, 0) {NLL};
        \draw[baseline] (0,0) -- (7,0);
        \foreach \x in {2.84, 3.55, 5.11, 5.66} \node[de_point] at (\x, 0) {};
        \node[en_point] at (3.10, 0) {};
        \node[cross, anchor=west] at (7.2, 0) {$\times$};
    \end{scope}

    \begin{scope}[yshift=1cm, xscale=0.1]
        \node[left, label_style] at (-6, 0) {BPC};
        \draw[baseline] (0,0) -- (84,0);
        \foreach \x in {40.9, 42.8, 22.5, 37.9} \node[de_point] at (\x, 0) {};
        \node[en_point] at (64.6, 0) {};
        \node[check, anchor=west] at (86, 0) {\checkmark};
    \end{scope}

    \begin{scope}[yshift=0cm, xscale=0.1]
        \node[left, label_style] at (-6, 0) {BPC};
        \draw[baseline] (0,0) -- (84,0);
        \foreach \x in {39.1, 30.4, 14.5, 38.3} \node[de_point] at (\x, 0) {};
        \node[en_point] at (36.8, 0) {};
        \node[cross, anchor=west] at (86, 0) {$\times$};
    \end{scope}

    \begin{scope}[shift={(0.0, -1)}]
        \node[de_point, label=right:{$\de_1-\de_4$}] at (0, 0) {};
        \node[en_point, label=right:{$\en$}] at (3.2, 0) {};
        \node[anchor=west] at (5, 0) {\textcolor{inconsistent}{$\times$} / \textcolor{consistent}{\checkmark} (In)consistent};
    \end{scope}

\end{tikzpicture}}
    \caption{Each line represents a row in a parallel dataset, each dot is an individual sentence; four German and one English. Common multilingual evaluations center around comparing intrinsic metrics such as the negative log-likelihood \mbox{(NLL)} or bits per character \mbox{(BPC)} with the assumption that these comparisons are fair since the \emph{semantic meaning} is consistent. However, these metrics measure \emph{information content} in the information-theoretic sense. This results in (1) differences within a language ($\de_i\leftrightarrow\de_k$), and (2) inconsistency across languages: if the EN sentence falls outside the range of the DE sentences it is \textcolor{consistent}{consistent}. If it falls within the range it is \textcolor{inconsistent}{inconsistent}, meaning conclusions can flip depending on the DE sentence we choose in our test set.}
    \label{fig:overview}\vspace{-1em}
\end{figure}
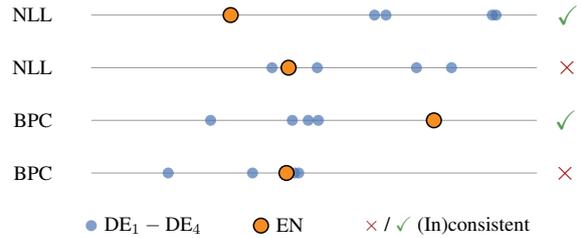

A common approach in both setups is to calculate the loss (or transformations of it) on multi-parallel datasets \cite[\eg][]{cotterell2018are,gerz2018language,mielke2019what,park2021morphology,wan2022fairness,chang2024when,chang2024goldfish,limisiewicz2024mytea,arnett2025why}.
A popular choice is the FLORES-200 machine translation dataset \cite{nllbteam2022no}.
Multi-parallel datasets are presumed to allow for a fair evaluation: samples have the same \emph{semantic meaning} in all languages.
However, the fairness of comparing metrics using multi-parallel datasets is less clear.
CLMs are optimized using an \emph{information-theoretic} objective, namely to minimize the cross-entropy loss, which is explicitly not about semantics.
We make such assumptions explicit and ask:

\begin{enumerate}[itemsep=1pt]
    \item How does the distinction between \emph{semantic meaning} and \emph{information content} apply to intrinsic evaluation metrics?
    \item What are the factors at play when comparing metrics, within and across languages?
    \item What are assumptions about comparing metrics when using a single multilingual model or multiple monolingual models?
\end{enumerate}
To answer these questions, we first discuss six existing intrinsic metrics and show how they relate to each other.
Next, we formalize the problem of comparing intrinsic metrics and relate it to the form-meaning debate.
Third, we look at the behavior of the metrics on two parallel corpora for mono- and multilingual models.
Fourth, we perform experiments with paraphrases to test the \emph{consistency} of the metrics (Figure~\ref{fig:overview}).
Finally, we take a broader view and tie our findings to the larger question of intrinsic CLM evaluations, the form-meaning debate, and linguistic information content.

\section{Related Work}\label{sec:related}
\paragraph{Multilingual Language Modeling.}
A CLM ($\mathcal{M}$) is trained to model the distribution of a token $w_i$ conditioned on its preceding context:
\begin{equation}\label{eq:1}
    \text{NLL}(\mathcal{M}) = -\sum_{i=1}^{S} \log P(w_i | w_1, \dots, w_{i-1}; \mathcal{M}).
\end{equation}
The model has to maximize the log-probability, or equivalently, minimize the negative log-likelihood (NLL), of the cross-entropy of the model's distribution and the true distribution of the next token.

As mentioned, multilinguality for CLMs can be tackled from two angles: a multilingual model that includes many languages \cite[\eg][]{scao2022bloom,ustun2024aya,chang2024when} or comparing multiple monolingual models \cite[\eg][]{mielke2019what,park2021morphology,chang2024goldfish}.
Our study is closer to the second, although we discuss implications for both.

\paragraph{Intrinsic Evaluation.}
Intrinsic metrics for CLMs are based on how well a model predicts tokens in an unseen test sequence; perplexity being the most common.
In order to make the evaluation fair, multi-parallel data is used.
However, this type of data can have issues, such as translation effects or other artifacts.
This has resulted in a number of slight variations or transformations of NLL that attempt to address these issues \cite[\eg][]{cotterell2018are,wan2022fairness,tsvetkov2024information}.
We discuss the metrics in detail in \S\ref{sec:metrics}.

\paragraph{Consistency.}
The ability of language models to equivalently handle paraphrases is known as \emph{consistency} \cite[\eg][]{elazar2021measuring,ohmer2024forms}.
Recently, \citet{poelman2025confounding} noted that comparing intrinsic metrics in multilingual settings could be unreliable due to being dependent on the form of parallel data.
We repeat their hypothetical example in Table~\ref{tab:ppl}.
\begin{table}[h!]
	\centering
	\small
	\resizebox{0.85\linewidth}{!}{\begin{tabular}{llc}
    \toprule
    Model & Sequence & PPL \\
    \midrule
    A & Sabe  jugar  al  ajedrez & 20\\
    B & Do  you  know  how  to  play  chess & 22 \\
    B & Can  you  play  chess & 18 \\
    \bottomrule
\end{tabular}
}
	\caption{Two valid parallel English sentences and a Spanish sentence with hypothetical PPLs. If we select the first parallel English option, model A is ``better'' than B, and vice-versa for the second.}
    \vspace{-1.5em}\label{tab:ppl}
\end{table}

\paragraph{Linguistic Universals.} 
Instead of intrinsic metrics, we can also sample text from a CLM to evaluate how close it is to human language.
Human languages seem to have a distinct ``statistical fingerprint''.\footnote{We refer to \citet{bentz2023zipfian} for recent empirical work and to \citet{sproat2023symbols} for a broader overview.}
Perhaps the most famous metric related to this fingerprint is \citeauthor{zipf1949human}'s law (\citeyear{zipf1949human}): words in a corpus are inversely proportional to their frequency rank. This has been shown to hold across a number of human languages \cite{piantadosi2014zipfs}.
The type-token ratio \cite[or Heaps' law;][]{herdan1960typetoken} is another law that characterizes vocabulary growth.
Since the true distribution of human language is not known outside of careful experiments \cite{jumelet2023transparency}, we have to evaluate models by using test sets or by sampling text from models.

In the context of CLMs, \citet{meister2021language} measure how well generated output from models obeys Zipf's law and other statistical tendencies for English.
\citet{takahashi2017neural,takahashi2019evaluating} perform similar studies for English and Chinese.
However, the usefulness of linguistic laws when evaluating CLMs is debatable.
Zipf's law can be found in many unexpected places \cite{piantadosi2014zipfs}.
Similarly, a model can produce gibberish while fully adhering to either law.
For these reasons, we choose to focus on intrinsic evaluation metrics. 
We come back to the laws in \S\ref{sec:discussion}.

\section{Metrics}

\subsection{Intrinsic Metrics}\label{sec:metrics}
We define an ``intrinsic metric'' as measuring how well a CLM predicts an unseen test sequence.
In Eq.~\ref{eq:1}, we define how a CLM is trained.
Given a trained CLM, we can give it a sequence ($S$) and get the average loss (NLL; $\downarrow$) across the tokens:
\begin{equation}
   \text{NLL}(S)
   = -\frac{1}{T}
   \sum_{t=1}^{T}
   \log P(w_t \mid w_{<t}).
\end{equation}
Lower values mean that the model assigns higher probabilities to
the observed tokens. \emph{Perplexity (PPL; $\downarrow$)} is a
monotonic exponential transformation of NLL:
\begin{equation}
   \text{PPL}(S) = \exp\left(\text{NLL}(S)\right).
\end{equation}
NLL can also be expressed as a total coding cost. Let $I(S)$ denote
the total coding cost of $S$ in bits:
\begin{equation}
   I(S)
   = -\sum_{t=1}^{T}
   \log_2 P(w_t \mid w_{<t}).
\end{equation}
If $|S|_{\mathrm{char}}$ is the length of $S$ in characters,
\emph{Bits per Character (BPC; $\downarrow$)} is
\begin{equation}
   \text{BPC}(S)
   = \frac{I(S)}{|S|_{\mathrm{char}}}.
\end{equation}
BPC measures the average number of bits required to encode each
character. Given target-language text $S_L$ and an aligned English
counterpart $S_{\mathrm{EN}}$, \emph{Bits per English Character}
\mbox{(BPEC; $\downarrow$; \citealp{cotterell2018are})} and
\emph{Information Parity}
\mbox{(IP; $\uparrow$; \citealp{tsvetkov2024information})} are
\begin{equation}
   \text{BPEC}(S_L)
   = \frac{I(S_L)}
          {|S_{\mathrm{EN}}|_{\mathrm{char}}},
   \quad
   \text{IP}(S_L)
   = \frac{I(S_{\mathrm{EN}})}
          {I(S_L)}.
\end{equation}
For corpus-level evaluation, coding costs and character counts are
summed over all sequences before calculating the metrics.
Finally, \emph{Mean Reciprocal Rank} \cite[\emph{MRR; $\uparrow$;}][]{limisiewicz2023tokenization} evaluates the model's \emph{ranking} of the true token:
\begin{equation}
    \text{MRR}(S) = \frac{1}{S} \sum_{t=1}^S \frac{1}{R_t},
\end{equation}
where $ R_t $ is the rank of the correct token in the model's predicted distribution over the vocabulary $V$ (with size $|V|$). 
Higher MRR means the model places the correct token near the top of its predictions.
Intuitively, MRR is a more lenient metric than the NLL-based metrics since it does not matter which probability the model assigns: MRR stays the same if $w_t$ is at the top rank with 0.9 or 0.02.

\subsection{Metric Usage}\label{sec:usage}
As mentioned, intrinsic metrics are often used with multi-parallel datasets.
This keeps the semantic meaning constant across languages.
There are a number of reasons for using the metrics in this way.

\citet{park2021morphology} mention that \emph{``because each [sequence] is intended to express the same meaning across languages, differences in (\ldots) surprisal primarily indicate differences in cross-linguistic language model quality.''}

Regarding potential translation issues, \citet{mielke2019what} outline that \emph{``different surprisals on the translations of the same sentence reflect quality differences in the language models, unless the translators added or removed information. (\ldots) if we find NLL(A) $>$ NLL(B), we must assume A contains more information, or that our language model was simply able to predict it less well.''}

\citet{tsvetkov2024information} discuss the effect of compression: \emph{[IP] measures how efficiently the LLM represents information provided by a text in the language L compared to the same information provided in English. A higher IP indicates a higher representation efficiency hence a closer alignment with the ideal language-agnostic compressor.}

Compression is also considered by \citet{wan2022fairness}: \emph{[We use] the total number of bits needed to encode the [parallel] dev set [BPC without averaging]. [This is a] more general and flexible way of evaluating data that has not been or cannot be perfectly segmented or aligned line by line.}

These are all valid concerns. 
However, all metrics use NLL and transform it in some way.
Since the dataset is multi-parallel (consistent meaning), a common assumption is that metrics are therefore directly fair to compare.
However, this may not hold: consistent meaning does not ``neutralize'' different forms with the same meaning.
The core of training a CLM is an information-theoretic objective, which is not about \emph{meaning content} as we know it from semantics, it is instead about \emph{information content}: the probability of an event occurring (for CLMs, we can say predicting a token is the event, with the vocabulary of the model as the options).
Information content in this sense is explicitly not about semantics \cite{shannon1948mathematical,shannon1951prediction}.
We outline this problem in more detail in the next section.

\section{Theoretical Outline}\label{sec:comparing}
In order to compare the metrics, we first need to know what factors are involved.
We outline the tokenization, dataset, and modeling setups we consider.
Let $\langs = \{\lang_1, \lang_2, \dots, \lang_N\}$ be a set of $N$ languages in our experiment (and $\langs^* \cup \langs = \langs^\dagger $ are ``all human languages'').
For each language $\lang_i$, we have a \emph{Monolingual Model} ($\mmono_i$), which uses a \emph{Monolingual Tokenizer} ($\tmono_i$), both trained on a \emph{Monolingual Corpus} ($ \cmono_i $).
The \emph{Multilingual Model} ($ \mathcal{M}^{\text{Multi}} $) and \emph{Multilingual Tokenizer} ($ \tmulti $) are trained on \mbox{$ \cmono_i \in \cmulti $}.
We consider $\cmulti$ to be a \emph{multi-parallel} corpus: \mbox{$ \cmono_{i,j} \leftrightarrow \cmono_{k,j} $} are parallel translations of the sample $j$ for $\lang_i$ and $\lang_k$.
Multi-parallel means $|\cmono_{i}| = |\cmono_{k}|$ for any pair in $\langs$ and that the rows are all consistent in their meaning.
We indicate the test split of a corpus as \emph{corpus-prime}: $ \cmonotest $.
Each tokenizer-model combination has a vocabulary $\vocab$.
When evaluating any model $\model$, we assume any of the metrics from \S\ref{sec:metrics} are used; for our running example we use NLL.

\paragraph{Monolingual.}
Inherently, the corpora $\cmono_i$ and $\cmono_k$ are not the same.
This means the models $\mmono_i \neq \mmono_k$, regardless of the choice of segmentation.
For example, if two languages share an alphabet and we train character-based models,\footnote{So assuming no training is necessary for the tokenizers.} we end up with $\tmono_i = \tmono_k$ and $\vocab_i = \vocab_k$.
Since the corpora are not the same, this means the distribution learned by the models will be different.
Often times, segmentations and vocabularies are not the same between languages, leading to $\tmono_i \neq \tmono_k$.
Still, because $\cmono_i \neq \cmono_k$, we can say for the respective models $\mmono_i$ and $\mmono_k$ that the distribution they learn is not the same, even if the sequences express the same meaning.
All metrics outlined in \S\ref{sec:metrics} are affected by this in the monolingual setting.
This means different distributions are being used and compared, both in training and testing.
Whether this is desirable comes down to the question of the existence of a universal statistical fingerprint to any human language (see \S\ref{sec:related} and \S\ref{sec:discussion}), whether this distribution can be accessed by any single language, and whether the metrics are the right tool for this.

\paragraph{Multilingual.}
Describing the distributions for $ \mmulti $ and $ \tmulti $ is more nuanced.
Starting from the corpus $\cmulti$, we can identify how many tokens are overlapping between $\cmono_i$ and $\cmono_k$ (or, $\vocab_i$ and $\vocab_k$), regardless of the segmentation choice.
For overlapping tokens, we explicitly combine the probability distributions of $\lang_i$ and $\lang_k$ in $\mmulti$.
When doing this for all languages ($\langs$), we essentially ``merge'' (parts of) their distributions.
Even when disallowing overlapping tokens in $\vocab^{\text{Multi}}$, or assuming $\cmono_i$ and $\cmono_k$ share no tokens, it is often assumed a multilingual model can learn from this combined or ``universal''\footnote{As mentioned, the true distribution of human language is not known. Previous research has referred to multilingual approaches as ``universal'' \cite[\eg][]{yang2020multilingual}.} distribution across languages.
This is the basis of \emph{zero-shot} language modeling, where a multilingual model can process one or more $\lang^* \in \langs^*$.
In practice, there are many factors involved: writing systems, (cross-lingual) homographs, polysemy, the aforementioned allocation of $\langs$ in $\vocab^{\text{Multi}}$ or $\tmulti$, and more.

At least in principle, we can assume $\mmulti$ is able to process\footnote{There are many caveats regarding tokenization and the handling of out-of-vocabulary items. We take a broad view since we are mainly talking about the metrics.} tokens of $\langs$.
So, with some caveats, we can assume:
\begin{equation}
    P(\langs) \approx P(\langs^\dagger).
\end{equation}
This means metrics from \S\ref{sec:metrics} can, to some extent, be meaningfully compared between any pair in $\langs$ since the model and tokenizer both have seen the distributions through $\cmulti$.

\paragraph{Paraphrases.}\label{sec:para}
As mentioned, a multi-parallel corpus removes confounds such as corpus size imbalances between languages and ensures consistent meaning across languages.
We now ask the question ``are all metrics comparable across all languages when we use the multilingual components we have discussed?''
For this, we look at the \emph{consistency} of intrinsic metrics across paraphrases.
We define $ \calttest_{k} $ as alternatives or paraphrases  of $\cmonotest_k$ for $ \lang_k $.
If we take a parallel sequence from either set and if the metrics are comparable \emph{and} if they measure a ``universal'' distribution (either in meaning, or in information content), we should see:
\begin{equation}
    \text{NLL}(\cmonotest_{k,j}) \approx \text{NLL}(\calttest_{k,j}).
\end{equation}
This is \emph{within} a language $\lang_k$.
There is also the question of comparing \emph{across} languages.
A strict interpretation of consistency across languages is that the NLL of parallel sequences has to be similar for any two $\lang_i$ and $\lang_k$:
\begin{equation}
    \text{NLL}(\cmonotest_{i,j}) \approx \text{NLL}(\cmonotest_{k,j}),
\end{equation}
but as noted before by several works in \S\ref{sec:usage}, this adds factors: how good is the language model? Is one language harder to model than another?

We can loosen this interpretation by taking inspiration from MRR and measure the relative \emph{ranking} of sequences between \mbox{$\cmonotest_{i,j} \leftrightarrow \cmonotest_{k,j} \leftrightarrow \calttest_{k,j}$}.
Differences in individual values due to the causes listed above should not affect the ranking if the metric is at least consistent.
For example, assume we have one English sentence and four German sentences, all with the same meaning.
If the values for an intrinsic metric for the four German sentences are \emph{all lower} or \emph{all higher} than the English sentence, the ranking of English and German is consistent for that sample.
We can test if this holds per sample in the test data (Figure~\ref{fig:overview}).
We can also average across the splits (so $\overline{\cmonotest_k}$ and $ \overline{\calttest_k}$) and see if those are consistent when comparing with $\overline{\cmonotest_i}$.
This ranking of averages is the most lenient interpretation of consistency.
Coming back to our example: it allows room for German being potentially harder to model or a model being worse at modeling German, as well as some inconsistencies from sample to sample.
If the ranking on the level of samples or averages turns out to be inconsistent (e.g., two German sentences score lower and two score higher compared to the English sentence, or the average of the English set is in between two German sets), the interpretation of any comparison of the metric is ambiguous and the question arises of what we are comparing and if this comparison is meaningful.

\paragraph{Recap.}
To summarize, we outline three potential issues with the intrinsic metrics:
\begin{enumerate}
    \item We are inherently comparing different distributions with the metrics for any $\mmono_i$ and $\mmono_k$, or we have to assume there is a universal distribution to all human languages which we can access using a metric through a single language: $L \in \langs^\dagger$.
    \item We are comparing a shared distribution when using $\mmulti$ since it has seen the combined distributions of $\langs$. If we take zero-shot capabilities of $\mmulti$ into account, we can more reliably assume we are evaluating some idea of a universal distribution: the larger or the more diverse $\langs$ gets,\footnote{Multilingual generalizability comes either from a \emph{large} set of languages, or from a \emph{representative} set \cite{bender2011achieving,ploeger2024what}.} the closer we get to modeling $\langs^\dagger$.
    \item If we assume the metrics consistently measure semantic meaning of parallel sequences (or are impervious to it), we should see roughly the same values for paraphrases within a single language. Similarly, across languages we should also see roughly the same values in a strict interpretation. With a less strict interpretation, the ranking of \mbox{$\cmonotest_{i,j} \leftrightarrow \cmonotest_{k,j} \leftrightarrow \calttest_{k,j}$} should stay consistent, either on a sample-level (Figure~\ref{fig:overview}) or when using averages across splits (Table~\ref{tab:subsets}).
\end{enumerate}

\section{Method}
We perform the following experiments:
(1) We train $\mmono$ and $\mmulti$ models and their tokenizers to study the behavior of the metrics.
(2) We test paraphrase consistency as outlined in \S\ref{sec:para}.

\paragraph{Data.}
All corpora we use are multi-parallel ($\cmulti$).
For training tokenizers and models, we use EuroParl \cite{koehn2005europarl} and
the UN Parallel Corpus \cite[UNPC,][]{ziemski2016united}.
We train $\mmono$ and $\mmulti$ variants.
EuroParl covers 21 languages and UNPC six.
The sizes of the datasets are smaller than what is commonly used to train state-of-the-art LLMs.
However, we specifically need to train and evaluate on multi-parallel data for our experiments, which limits us in our choice of datasets.
Full details are in \S\ref{app:datasets}.

For evaluation, we use the aforementioned FLORES-200 dataset \cite{nllbteam2022no}.
For testing paraphrase consistency, we use translation data from \citet{freitag2020humanparaphrased,freitag2020bleu}, which we discuss further in \S\ref{sec:para-results}.
\begin{figure*}[ht]
    \centering
    \includegraphics[width=\linewidth]{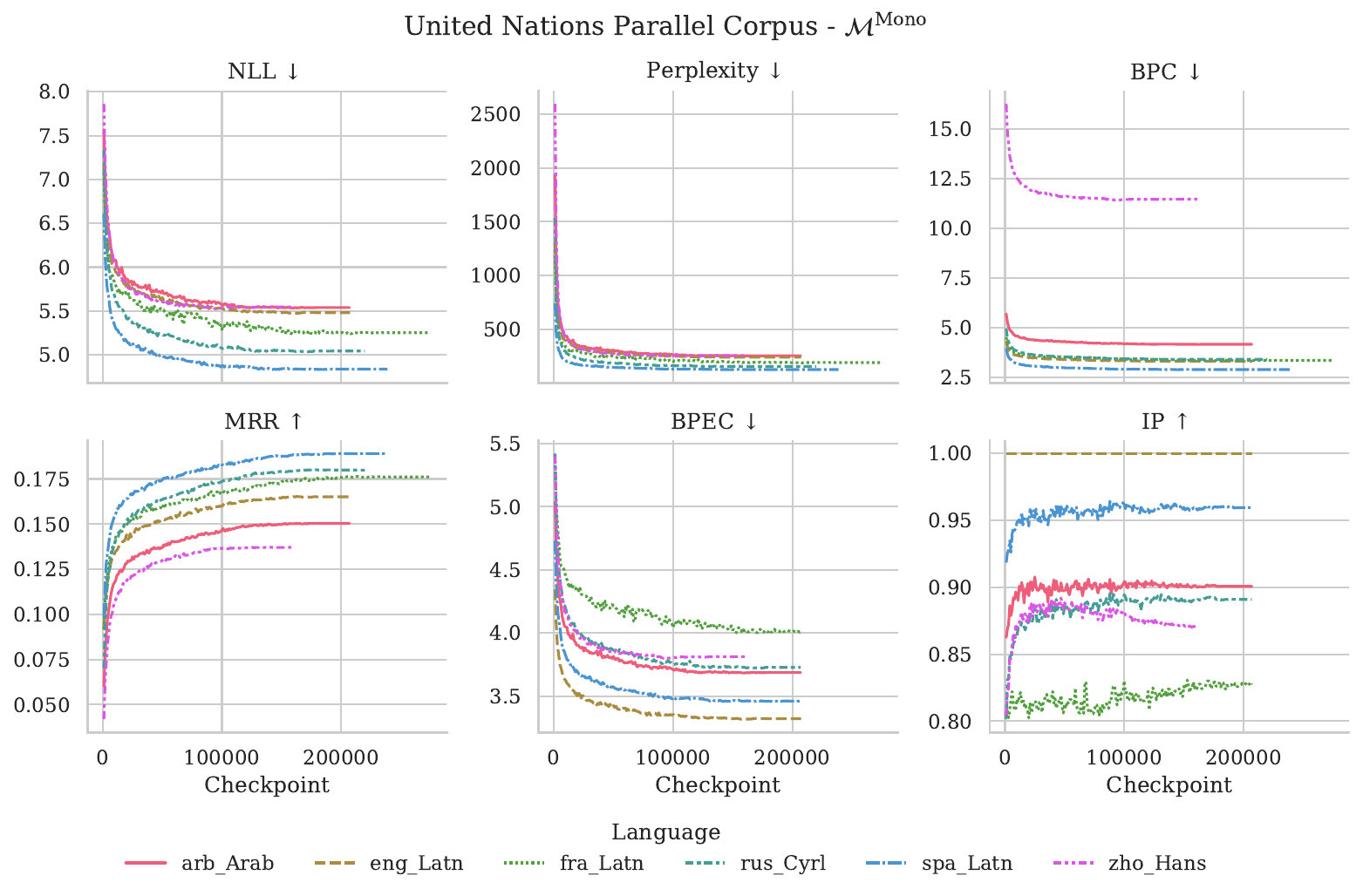}
    \caption{Metrics across checkpoints during training when evaluated on FLORES-200. The different lengths of the lines is due to tokenization differences between languages, resulting in shorter or longer sequences. The models have seen the same amount of parallel data when measured in number of lines.}
    \label{fig:lm-un-mono}\vspace{-1em}
\end{figure*}

\paragraph{Models and Tokenization.}
We aim to keep the tokenization and model choices as ``mainstream'' as possible.
We train $\tmono$ and $\tmulti$ variants on the same training data as the models.
We use BPE \cite{gage1994new,sennrich2016neural} with byte-level fallback.
All $\tmono$ receive a vocabulary size of 32k and $\tmulti$ 150k, based on previous estimates \cite[\eg][]{conneau2020unsupervised,xue2021mt5,ustun2024aya}.
We train a number of tiny Llama-3-style \cite{grattafiori2024llama} models using Huggingface \texttt{transformers} \cite{wolf2020transformers}.
We apply insights about training tiny models from the MobileLLM project \cite{liu2024mobilellm}. 
Small models are sometimes deemed unreliable since some phenomena only appear at larger scales.
However, small models are suitable for specific investigations \cite[\eg][]{chang2024when,tatariya2025how,wilcox2025bigger}, such as ours.
While the absolute values for the intrinsic metrics might change, the potential inconsistency of the comparisons will not suddenly disappear at larger scales.
Additionally, large models do not add much when our datasets are this small.
Full training details are listed in \S\ref{app:parameters}.

\paragraph{Metrics.}
The metrics are calculated for all sentences in FLORES-200 for a specific language for $\mmono$ and for all supported languages ($\langs$) for $\mmulti$. 
We do this every 1000 steps.
BPEC and IP require the coding costs of the target and aligned English texts. For $\mmono$, we obtain this cost from $\mmono_{\mathrm{EN}}$. 
For $\mmulti$, both costs are obtained from the same multilingual model.

\section{Results}\label{sec:results}
Figure~\ref{fig:lm-un-mono} shows the behavior of the metrics when using monolingual models trained on UNPC.
We can see English, Arabic, and Chinese have similar NLL values, all being quite high (lower is better).
Perplexity is a transformation of NLL, so it does not show new rankings.
BPC shows Chinese as a clear outlier, likely because the content is expressed using substantially fewer characters compared to the other languages. This difference is reduced under BPEC and IP, which account for the text lengths and coding costs.
However, French is a clear outlier here.
This is caused by its translations being quite lengthy.
Can we conclude that French is harder to model? Even when its NLL is \emph{lower} than English?
We go into detail about why these intrinsic metrics might not be suitable for such questions in the next section.
MRR shows a similar, but more consistent ranking as NLL (albeit inverse; higher is better).
Whereas Arabic, Chinese, and English trade places for the NLL-derived metrics throughout training, they remain quite consistent for MRR.

The multilingual results are similar (Figure~\ref{fig:lm-un-multi}), but the NLL is noticeably higher for all languages, likely due to the reasons listed in \S\ref{sec:usage}.
All languages using a Latin script are performing the ``best'' for $\mmulti$, whereas Russian is the second best for $\mmono$.
This could be due to the shared distribution outlined in \S\ref{sec:metrics}.
The question of whether we are evaluating how well this particular set of technical decisions is suited to model these languages or how difficult these languages are to model is not straightforward (see \S\ref{sec:discussion}).

Finally, the EuroParl results (\S\ref{app:lm-results}) are similar to the UNPC results, both for the $\mmono$ and $\mmulti$ settings.
One interesting aspect of EuroParl is that all 21 languages in the dataset except Bulgarian and Greek use the Latin script.
However, their values for the metrics do not stand out like Chinese or Arabic do for UNPC.

\begin{figure*}[ht!]
    \centering
    \includegraphics[width=\linewidth]{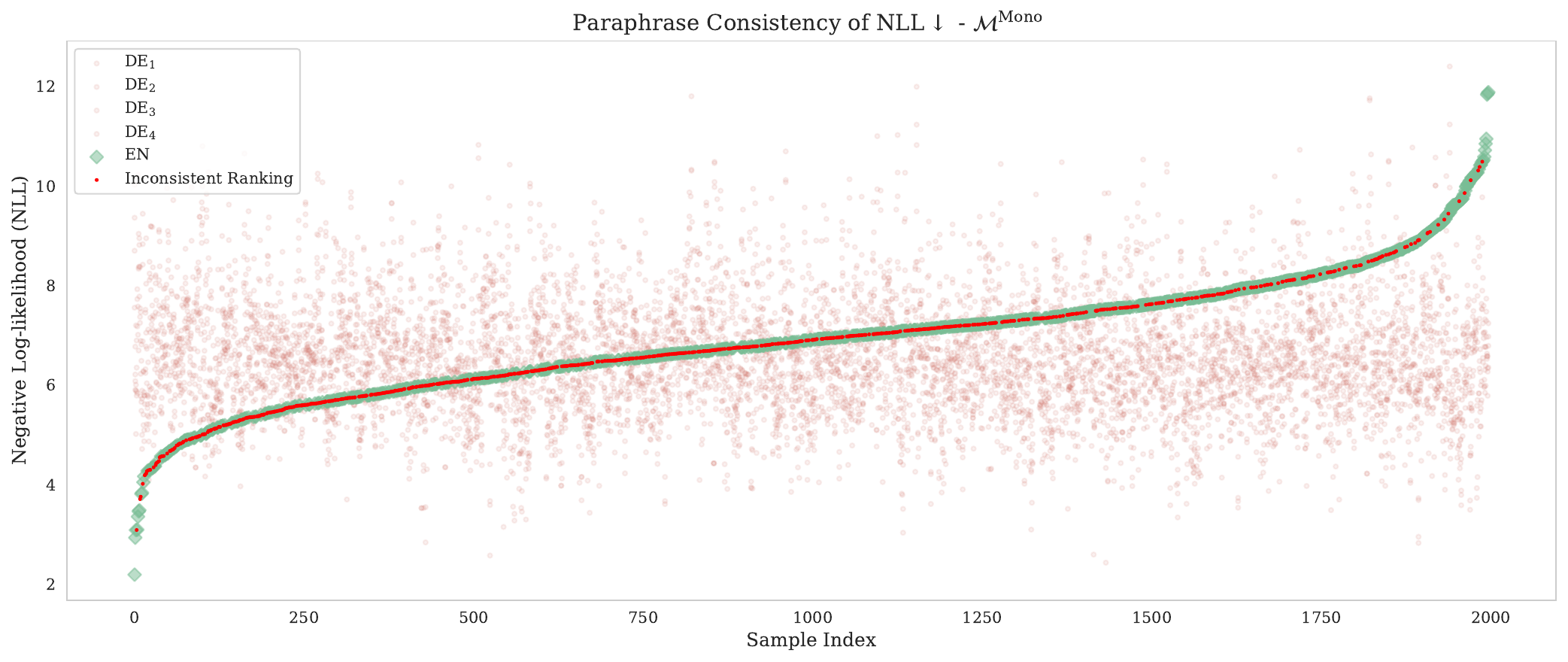}
    \caption{Paraphrase consistency of NLL values for the EN source and four parallel DE paraphrases using monolingual EuroParl models. On the x-axis, we list the sample index. Each sample consists of the EN sentence (green) and the four DE paraphrases (light red). For the sake of visual clarity, we sort the results by the English NLL. We show the ranking consistency: if all DE paraphrases are above or below the EN source, it means the ranking is consistent. Inconsistent rankings are marked with a red dot inside the green diamond for English. We see similar results for the multilingual model, as well as for the other metrics, see \S\ref{app:sensitivity-results}.}
    \label{fig:nll-sensitivity}\vspace{-1em}
\end{figure*}

\paragraph{Paraphrase Consistency.}\label{sec:para-results}
As shown in the previous section, it is tempting to draw conclusions about a language as a whole when looking at the results.
Regardless of whether the metrics measure information content or semantics, are they consistent when presented with paraphrases?
To test this, we use the \mbox{EN$\rightarrow$DE} set of \citet{freitag2020humanparaphrased,freitag2020bleu}.
Every source sequence has \emph{four} human translated references. 
This results in four parallel splits:
\begin{enumerate}
    \item The original reference translation ($\de_1$).
    \item An alternative reference translation ($\de_2$).
    \item A ``paraphrased as-much-as-possible'' version of the original reference ($\de_3$).
    \item A ``paraphrased as-much-as-possible'' version of the alternative reference ($\de_4$).
\end{enumerate}
All four German sentences and the English sentence convey the same semantic meaning.\footnote{The caveats mentioned in \S\ref{sec:usage} about adding or removing information in translations remain. Although this should not matter since we use this dataset in the exact same way FLORES-200 and other parallel datasets are commonly used.}
The dataset is similar in domain and size to FLORES-200: high quality news articles and about 2000 samples.
We use $\mmono_{\text{EN}}$ and $\mmono_{\text{DE}}$, as well as $\mmulti$ trained on EuroParl.
Additionally, we track the ranking of the source and four targets as a more lenient alternative to having similar NLL values.
Table~\ref{tab:consistency} summarizes the consistency of the metrics.
We can see that all metrics are highly inconsistent.
\begin{table}[h!]
    \centering
    \small
    \begin{tabular}{lcc}
    \toprule
     Metric & $\mmono$ & $\mmulti$\\
     \midrule
     NLL & 47\%  & 51\% \\
     BPC & 50\% & 52\%\\
     MRR & 50\% & 51\%\\
     \bottomrule
\end{tabular}

    \caption{Sample-level inconsistency. For instance: 47\% of all samples are inconsistent for NLL with $\mmono$. Figure \ref{fig:overview} shows an intuitive explanation.}
    \label{tab:consistency}\vspace{-0.3em}
\end{table}

Figure~\ref{fig:nll-sensitivity} shows the NLL consistency of $\mmono_{\text{EN}}$ and $\mmono_{\text{DE}}$ EuroParl models.
The two issues outlined before seem to be correct: the difference \emph{within} a language is notable (the light red dots should be close together vertically if the NLL values were similar), and the comparison \emph{across} languages is ranked consistently for only about half the samples.

A logical question now is \emph{does it average out?}
So far we kept the splits (columns) from \citet{freitag2020humanparaphrased,freitag2020bleu}, and with these splits the averages across $\de_1$, $\de_2$, $\de_3$, and $\de_4$ are consistent for all metrics.
However, since the \emph{rows} are all parallel in meaning (e.g., the sample with index 8 has the same meaning in all splits: \mbox{$\de_{1,8} \leftrightarrow \de_{2,8} \leftrightarrow \de_{3,8} \leftrightarrow \de_{4,8}$}), there is no reason we cannot make new splits, as long as we do not change the ordering of the rows.
If we sort the German samples row-wise across the four splits, we end up with ``easiest'' to ``hardest'' splits.
The averages of these new splits are \emph{not} consistent.
The original and sorted splits are shown in Table~\ref{tab:subsets}.

\begin{table}
    \centering
    \resizebox{\columnwidth}{!}{\begin{tabular}{ll|l|l|l}
    Model & Metric & \multicolumn{1}{c|}{$\de^{\text{F}}$} & \multicolumn{1}{c|}{EN} &  \multicolumn{1}{c}{$\de^{\text{S}}$}  \\
    \hline
    \multirow{3}{*}{$\mmono$} & $\nllarrow$  & 6.43--6.64 \consistent & 6.91 &  5.89--7.23 \inconsistent \\
                              & $\bpcarrow$  & 2.09--2.20 \consistent & 2.27 &  1.82--2.44 \inconsistent\\
                              & $\mrrarrow$  & 19.6--21.2 \consistent & 18.6 &  14.6--26.0 \inconsistent  \\\hline
    \multirow{3}{*}{$\mmulti$} & $\nllarrow$ & 6.81--7.02 \consistent & 7.05 &  6.25--7.60 \inconsistent \\
                              & $\bpcarrow$  &  2.40--2.51 \consistent & 2.54 &  2.13--2.77 \inconsistent\\
                              & $\mrrarrow$  &  22.5--23.7 \consistent & 21.4 &  17.4--28.7 \inconsistent \\
\end{tabular}
}
    \caption{Split-level averages. The original splits from \citet{freitag2020humanparaphrased,freitag2020bleu} are in $\de^\text{F}$ and the row-wise sorted splits in $\de^\text{S}$. We show the maximum and minimum of the four splits since this is enough to tell us whether the averages are consistent or not: if EN falls outside the DE split range, it is \textcolor{consistent}{consistent}, if it falls within, it is \textcolor{inconsistent}{inconsistent}. \emph{All} metrics are \textcolor{consistent}{consistent} for the $\de^\text{F}$ subset and \emph{all} become \textcolor{inconsistent}{inconsistent} for $\de^\text{S}$.}\vspace{-0.8em}
    \label{tab:subsets}
\end{table}
This has practical implications.
If we were to do a study on the difficulty of modeling German versus English, and pick the ``easiest'' split from the sorted splits, we would conclude that German is easier to model than English. If we were to pick the ``hardest'' split, we would arrive at the \emph{opposite} conclusion.
This seems to confirm the hypothetical example by \citet{poelman2025confounding} in Table \ref{tab:ppl}.

\section{Discussion}\label{sec:discussion}
Our theoretical analysis and empirical results raise questions about the comparability of intrinsic metrics in multilingual language modeling.
We cover some interpretations in this section.

\paragraph{Technical Decisions or Languages.}
An unsolved question in multilingual NLP is which set of technical decisions results in the ``best'' language model in terms of equal performance across languages.
On the one hand are the modeling decisions, ranging from tokenization to hyperparameters.
On the other hand are intrinsic differences between written languages, ranging from writing systems to corpus-specific characteristics.

Our findings (and similar studies) are limited by the technical decisions made and languages used.
It could be that there exists a set of technical decisions and languages (or even corpora) that \emph{do} show consistent values for the metrics and that \emph{are} robust towards paraphrases.
However, our requirement for multi-parallel corpora for both training and evaluation restricts us to the data and languages we used.
In this controlled setting the issue of consistency and comparability is apparent.
To summarize:
\begin{itemize}
    \item Model-to-model comparisons are fair if the same test set and the same segmentation are used. Different segmentations are potentially fair if scores are scaled to a shared unit such as characters \cite{mielke2019can,bauwens2024bitspercharacter}.
    \item Language-to-language comparisons come with a number of assumptions and issues. (1) Is there a ``universal'' human language distribution and can monolingual and multilingual models access it? If so, how reliable are the metrics to measure this? (2) Parallel semantic meaning in datasets does not ``neutralize'' the effect of form, both within and across languages. Intrinsic metrics are inherently sensitive to this. (3) Inconsistency of metrics can lead to opposite conclusions, depending on the choice of test samples or splits, even with multi-parallel datasets.
\end{itemize}

\paragraph{Form and Meaning.}
The distinction between form and meaning is the basis of a number of analyses in various fields \cite[\eg][]{frege1892uber,quine1960word,searle1980minds,harnad1990symbol}.
The ``octopus paper'' by \citet{bender2020climbing} is probably the most pertinent discussion of the topic in the context of language modeling.
Their main argument is that a language model inherently only learns \emph{form} and not \emph{meaning}.
In some way this is a revisit of the Chinese Room argument \cite{searle1980minds} with more focus on discourse, grounding, and intentionality.

We do not have to go that far, our findings and argument are purely about the usage and assumptions about intrinsic metrics in multilingual evaluations.
Similar empirical results have been found by \citet{ohmer2024forms} who find inconsistency in the answers of LLMs to paraphrased questions.
Our argument is based on the distinction of information content in the information-theoretic or semantic meaning sense. 
The metrics (and CLM training) operate on the former, even when we use data that is parallel for the latter.
Having parallel data does not neutralize different forms and the metrics, and thus the conclusions we draw based on them, are inherently sensitive to this.
However, as mentioned, it is certainly possible there exists a set of languages and technical decisions that result in consistent metrics.
This shows the metrics might be unsuitable for questions like \emph{which language is easier to model?} 
We may need to look elsewhere.
\citet{sproat2014database} propose an external approach:

\begin{quote}
    \vspace{-0.3em}
    \emph{``Which languages convey the most information in a given amount of space? This is a question often (\ldots) asked by engineers who have some information theoretic measure of `information' in mind, but rarely define how they would measure that information. If one had a database of close translations between a set of typologically diverse languages, with detailed marking of morphosyntactic and morphosemantic features, one could hope to quantify differences [in] information.''}
    \vspace{-0.3em}
\end{quote}
This database was unfortunately never fully completed it seems.
How we would use such a database for evaluating CLMs is not obvious, but issues regarding form, meaning, and language characteristics would largely be ``solved''.\footnote{Other cross-lingual meaning representations such as UMR or DRS could be a way to test this \cite{vangysel2021designing,abzianidze2017parallel}, but at the time of writing, these do not have the coverage of a dataset like FLORES-200.}

Another area to look for alternatives to the intrinsic metrics are the statistical linguistic laws.
Even though the ones listed in \S\ref{sec:related} might not be that useful, if we could formulate a law that accounts for paraphrases or semantics, it would solve the problem.
How to do this is another question entirely, one that we do not have an answer to.
Until then, the best we can do is to keep paraphrases and potential (in)consistencies in mind.

\section{Conclusion}
We investigate intrinsic evaluation metrics for multilingual conditional language models.
Such evaluations are often done with multi-parallel datasets, where the \emph{meaning} of samples is consistent across languages.
However, the metrics are designed to measure \emph{information content}.
We (1) introduce existing intrinsic metrics,
(2) formalize the problem of comparing the metrics in mono- and multilingual setups, (3) look at the behavior of the metrics on two parallel corpora for both setups, (4) perform experiments with \emph{paraphrases} to test the \emph{consistency} of the metrics, and (5) discuss what metric we might be looking for instead of existing ones.
We ultimately find that using intrinsic metrics to comparing languages requires some strong assumptions.

\section*{Limitations}

\paragraph{Generation Quality.}
Intrinsic metrics can be useful, but they are not perfect.
Lower perplexity values have been shown to sometimes correlate with more unnatural text \cite{kuribayashi2021lower}.
Scoring well on these metrics does not necessarily mean that a model is of higher quality.
We discuss the issue of \emph{comparability}, the question of \emph{quality} is outside the scope of our work.

\paragraph{Language Interactions.}
Language distributions are not as cleanly distinct as they are generally treated in NLP and (to some extent) our current study.
Distributions of many languages are inherently mixed due to loan words, colexification, homographs, code-mixing, and so on.
This is even more prevalent with web-based datasets that can contain ``unnatural'' language mixing due to problems with data quality \cite{blevins2022language} or with entirely new ways of mixing languages in prompt-based evaluations of language models \cite{poelman2025roles}.
Strictly dividing languages is necessary for furthering our understanding of language modeling, but caveats remain.

\ifx\review\undefined
\section*{Acknowledgements}
We thank François Yvon for pointing out a mistake in a previous version of this paper.
We thank Esther Ploeger for pointing us to the WMT paraphrase dataset, Thomas Bauwens for early discussions, and Coleman Haley for the valuable suggestion of looking into the question of \emph{does it average out?}
WP is funded by a KU Leuven Bijzonder Onderzoeksfonds C1 project with reference C14/23/096.
The computational resources and services used were provided by the VSC (Flemish Supercomputer Center), funded by the Research Foundation - Flanders (FWO) and the Flemish Government - department EWI.
\fi

\bibliography{custom}

\appendix
\onecolumn

\section{Datasets}\label{app:datasets}
\subsection{Pre-processing}
The training datasets (UNPC and EuroParl) are aligned in a multi-parallel way.
We first determine the language with the biggest overlap with all other languages; this is our pivot language (Arabic for UNPC and Italian for EuroParl).
Afterwards, we collect sequences that align with the pivot, resulting in our final datasets.
The test datasets (FLORES+ and WMT-19 paraphrases) are already multi-parallel for the languages we use, so did not require special pre-processing.
FLORES+ is a community effort to keep improving the FLORES test set.
We combine the \texttt{dev} and \texttt{devtest} splits from FLORES+ and use them as our test set.
Our development set consists of 10\% randomly sampled lines per language.
For monolingual models, this is just one language, for multilingual models, it depends on the languages included in the specific corpus.
Table~\ref{tab:datasets} describes the datasets and Table~\ref{tab:languages} lists their language coverage.

\begin{table*}[h]
    \centering
    \begin{tabular}{llrr}
    \toprule
    Dataset & Link & \multicolumn{1}{c}{$|\langs|$} & \multicolumn{1}{c}{$|\corpus|$} \\
    \midrule
    EuroParl \cite{koehn2005europarl} & \href{https://huggingface.co/datasets/Helsinki-NLP/europarl}{HF} & 21 & 211521 \\
    United Nations Parallel Corpus \cite{ziemski2016united} & \href{https://huggingface.co/datasets/Helsinki-NLP/un_pc}{HF} & 6 & 11290186  \\
    WMT-19 Paraphrases \cite{freitag2020humanparaphrased,freitag2020bleu} & \href{https://github.com/google/wmt19-paraphrased-references}{GH} & 2 & 1997  \\
    FLORES-200 (FLORES+) \cite{nllbteam2022no} & \href{https://huggingface.co/datasets/openlanguagedata/flores_plus}{HF} & 221$^*$ & 2009 \\
    \bottomrule
\end{tabular}

    \caption{Datasets used in our analyses. EP and UNPC are taken from the OPUS \cite{tiedemann2012parallel} collection on the Huggingface hub \cite{lhoest2021datasets}. The number of languages is listed, as well as the number of samples (rows) after multi-parallel alignment. $^*$The listed number is unique language-script combinations, not languages.}
    \label{tab:datasets}
\end{table*}

\begin{table}[h]
    \centering
    \begin{tabular}{lp{0.43\textwidth}}
     \toprule
     Dataset & Languages (ISO-639-3) \\
     \midrule
     EuroParl & bul, ces, dan, deu, ell, eng, est, fin, fra, hun, ita, lvs, lit, nld, pol, por, ron, slk, slv, spa, swe\\
     UNPC & arb, eng, fra, rus, spa, zho\\
     WMT-19 Paraphrases & deu, eng \\
     FLORES+ & Coverage for all languages in listed above.\\
     \bottomrule
\end{tabular}

    \caption{Language coverage of the datasets we use. Note that Latvian is listed as \texttt{lat} (Latin) in FLORES+, while it should be either \texttt{lav} (inclusive code) or \texttt{lvs} (Standard Latvian). We use the latter.}
    \label{tab:languages}
\end{table}

\clearpage

\section{Model Setup and Hyperparameters}\label{app:parameters}
Our code is available here: \href{https://github.com/WPoelman/form-and-meaning}{github.com/WPoelman/form-and-meaning}

\begin{table}[ht]
    \centering
	\begin{subtable}{0.5\textwidth}
        \begin{tabular}{ll}
\toprule
\textbf{Setting} & \textbf{Value} \\
\midrule
Architecture & LlamaForCausalLM \\
Attention Bias & False \\
BOS Token ID & $1$ \\
EOS Token ID & $2$ \\
Hidden Act & \texttt{silu} (swish) \\
Hidden Size & $576$ \\
Initializer Range & $0.02$ \\
Intermediate Size & $1536$ \\
Max Position Embeddings & $2048$ \\
Model Type & \texttt{llama} \\
Attention Heads & $9$ \\
Hidden Layers & $30$ \\
Key-Value Heads & $3$ \\
Pretraining TP & $1$ \\
RMS Norm $\epsilon$ & $1 \times 10^{-5}$ \\
ROPE Scaling & False \\
ROPE $\theta$ & $10000$ \\
Tie Word Embeddings & False \\
Torch Data Type & bfloat16 \\
\bottomrule
\end{tabular}

		\caption{Model architecture.}
	\end{subtable}
    \hfill
    \begin{subtable}{0.45\textwidth}
    \begin{tabular}{ll}
\toprule
\textbf{Parameter} & \textbf{Value} \\
\midrule
Epochs & $3$ \\
Learning Rate & $3 \times 10^{-4}$ \\
LR Warmup Steps & $2$ \\
LR Warmup Style & linear \\
LR Decay Style & cosine \\
Minimum Decay LR & $1 \times 10^{-5}$ \\
Zero Stage & $0$ \\
Weight Decay & $0.01$ \\
Clip Grad & $1.0$ \\
Accumulate Grad in FP32 & True \\
Optimizer & AdamW \\
Adam $\epsilon$ & $1 \times 10^{-8}$ \\
Adam $\beta_1$ & $0.9$ \\
Adam $\beta_2$ & $0.95$ \\
\bottomrule
\end{tabular}
\vspace{1em}

\begin{tabular}{lll}
\toprule
\textbf{Setting} & \textbf{$\mmono$} & \textbf{$\mmulti$} \\
\midrule
Vocabulary Size & 32k & 150k \\
Parameters & $\approx$ 143M & $\approx$ 279M \\
\bottomrule
\end{tabular}

        \caption{Training hyperparameters (top) and differences between monolingual and multilingual models (bottom).}
    \end{subtable}
    \caption{Model architecture and hyperparameters partially based on \citet{liu2024mobilellm}'s 125M Llama-style model.}
\end{table}

\begin{table}[ht]
    \centering
    \begin{tabular}{llp{0.5\linewidth}}
\toprule
\textbf{Step} & \textbf{Hardware} & \textbf{Cost (hours)}\\
\midrule
Dataset alignment & CPU & 10 total \\
Tokenizer training & CPU & 4 total \\
Monolingual training & H100 GPU & Less than 2 hours per model (27 models $\approx$ 50)\\
Multilingual training & H100 GPU & About 12 for UN-PC and 18 for EuroParl\\
\bottomrule
\end{tabular}

    \caption{Rough breakdown of compute used in the study. Estimates are somewhat exaggerated to cover experimentation and debugging.}
    \label{tab:compute}
\end{table}

\clearpage

\section{Full Results}
\subsection{Language Modeling Experiments}\label{app:lm-results}
\null\vfill
\begin{figure}[h]
    \centering
    \includegraphics[width=\linewidth]{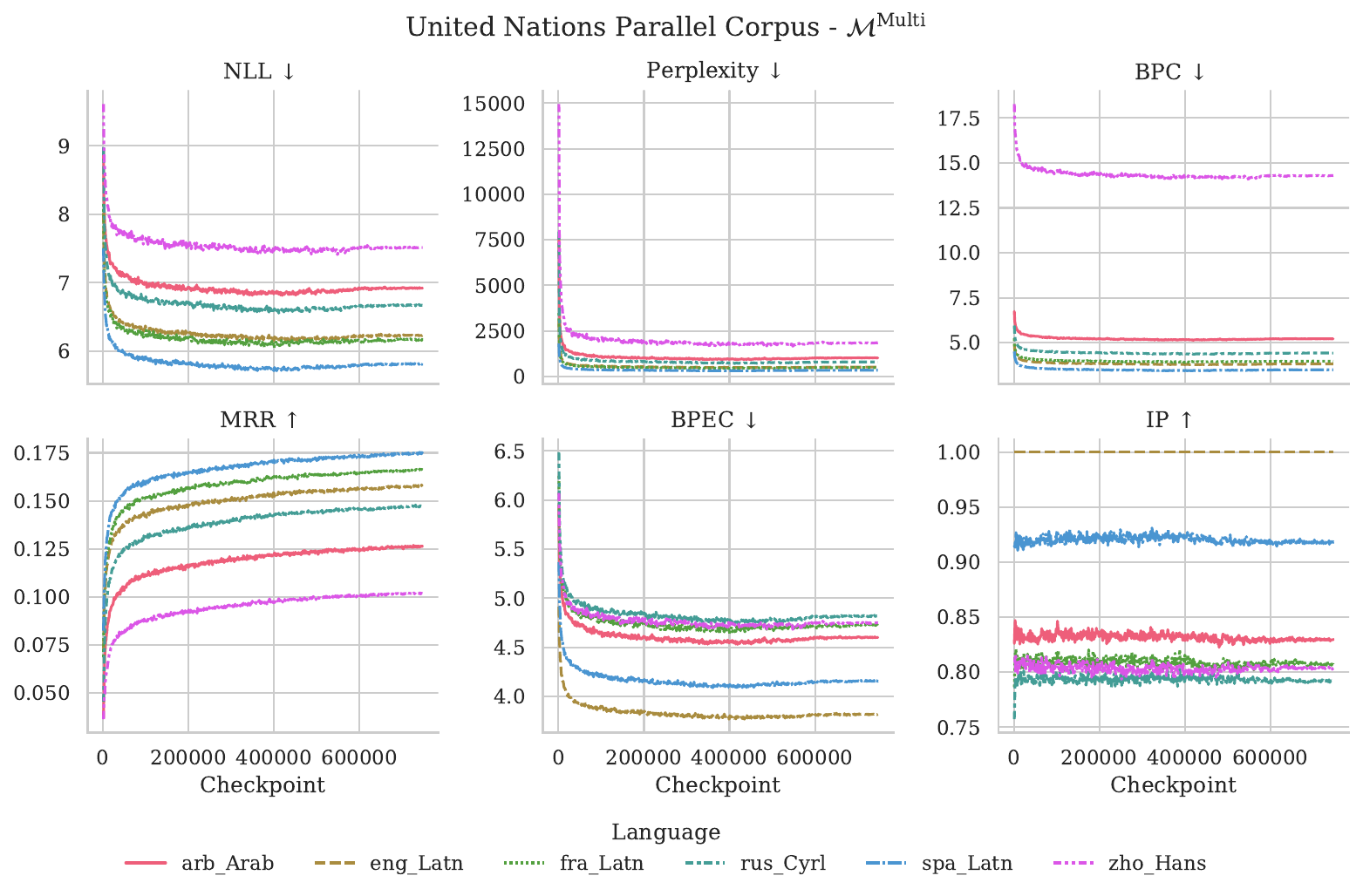}
    \caption{Metrics across checkpoints evaluated using FLORES-200 for the multilingual UNPC model.}
    \label{fig:lm-un-multi}
\end{figure}
\vfill\null

\begin{figure*}
    \centering
    \includegraphics[width=\linewidth]{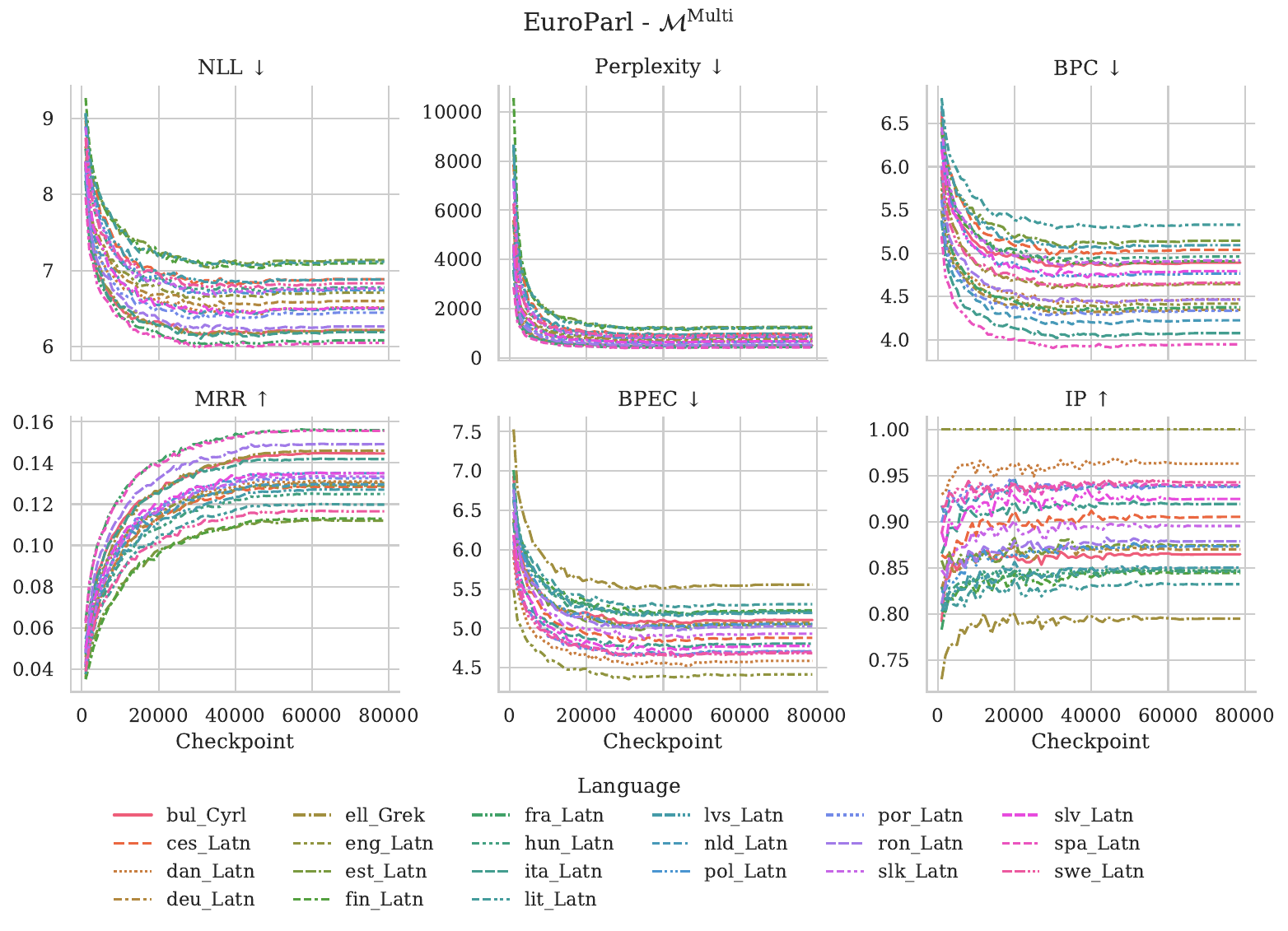}
    \caption{Metrics across checkpoints evaluated using FLORES-200 for the multilingual EP model.}
    \label{fig:lm-ep-multi}
\end{figure*}

\begin{figure*}[h]
    \centering
    \includegraphics[width=\linewidth]{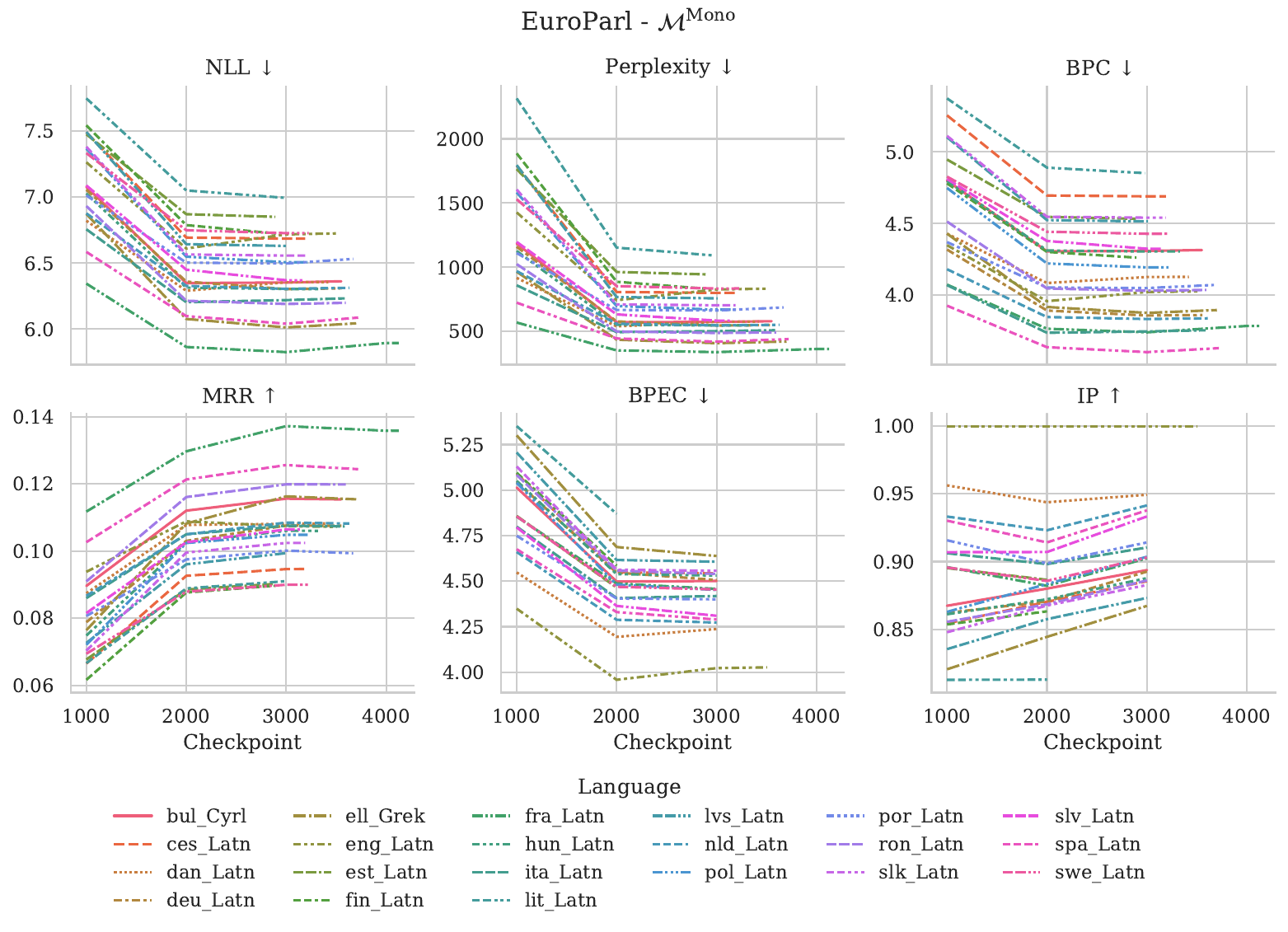}
    \caption{Metrics across checkpoints evaluated using FLORES-200 for the monolingual EuroParl models.}
    \label{fig:lm-ep-mono}
\end{figure*}

\clearpage

\subsection{Metric Consistency Experiments}\label{app:sensitivity-results}
\begin{figure*}[h!]
    \centering
    \includegraphics[width=\linewidth]{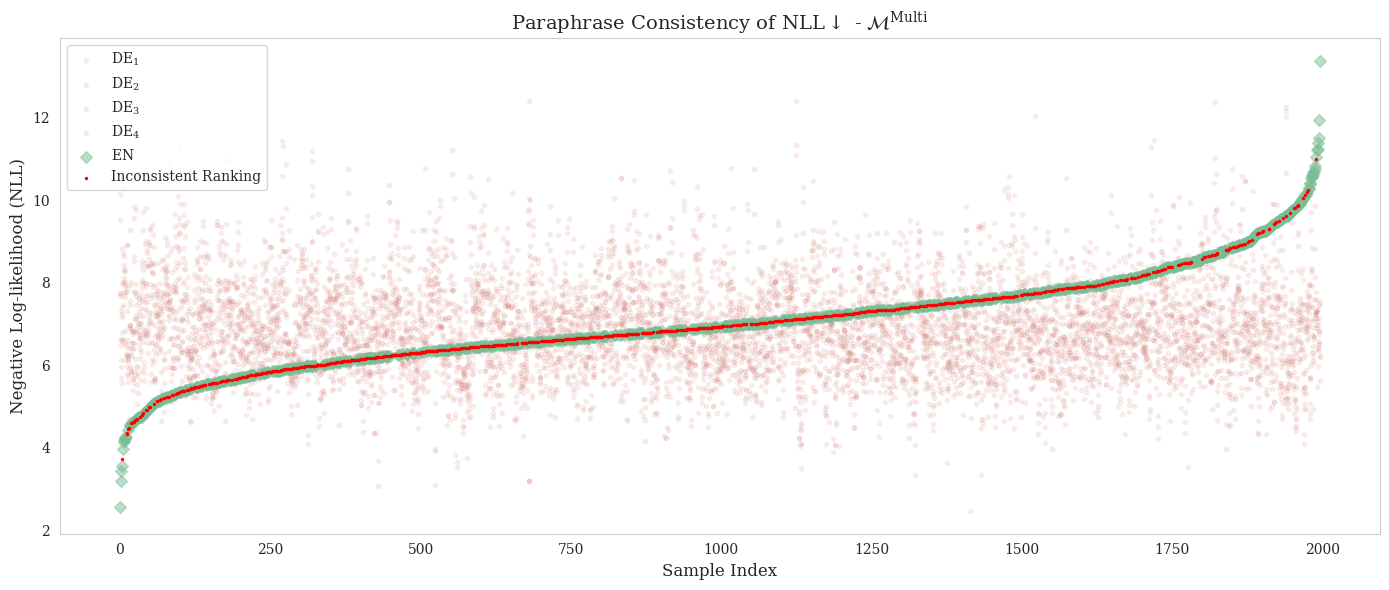}
    \includegraphics[width=\linewidth]{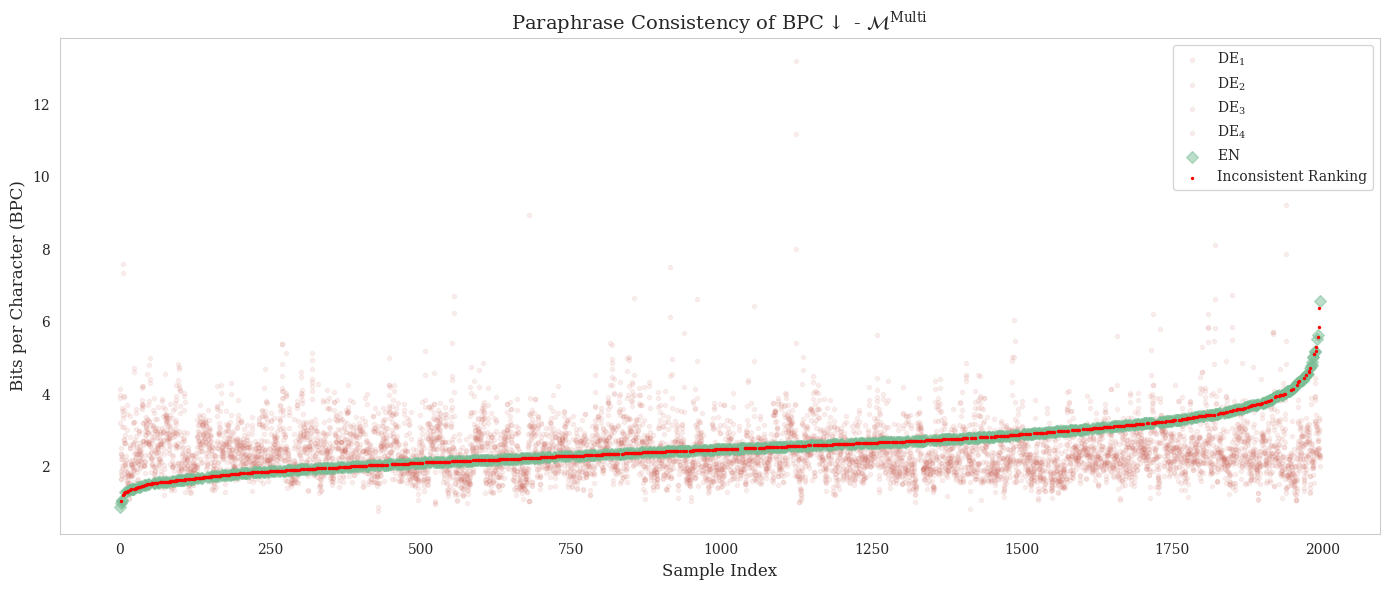}
    \includegraphics[width=\linewidth]{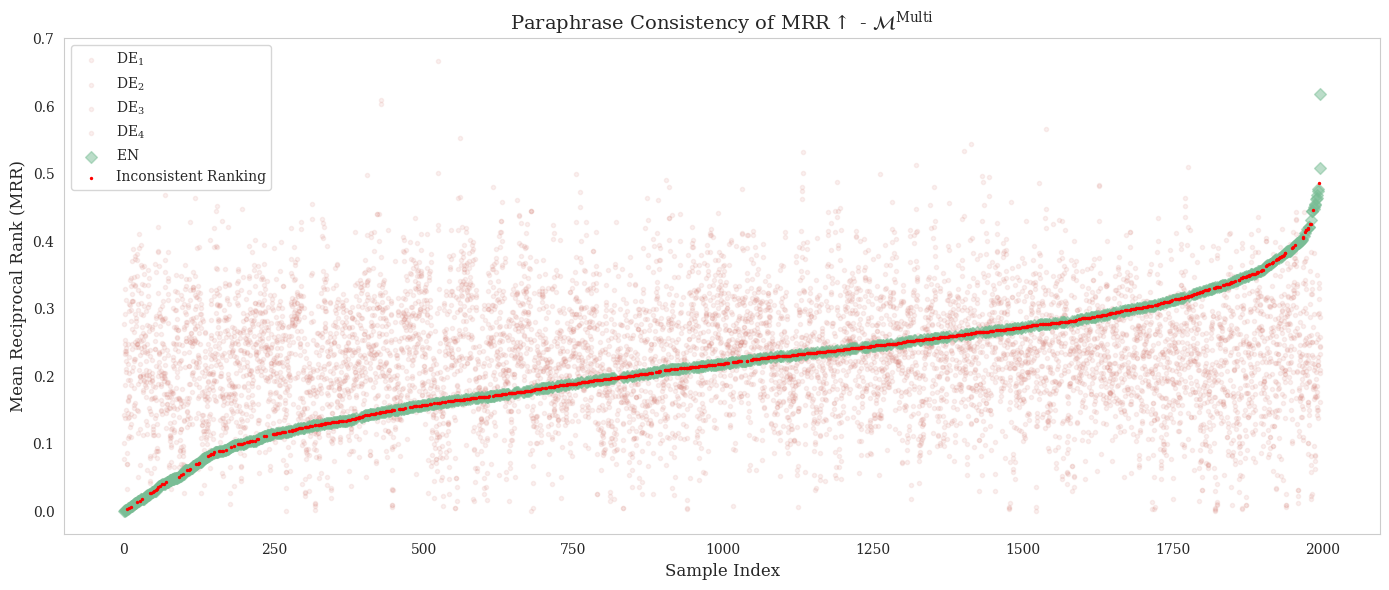}
    \caption{Sensitivity of NLL, BPC, and MRR using $\mmulti$ trained on EuroParl.}
    \label{fig:sens-ep-multi}
\end{figure*}

\begin{figure*}[ht]
    \centering
    \includegraphics[width=\linewidth]{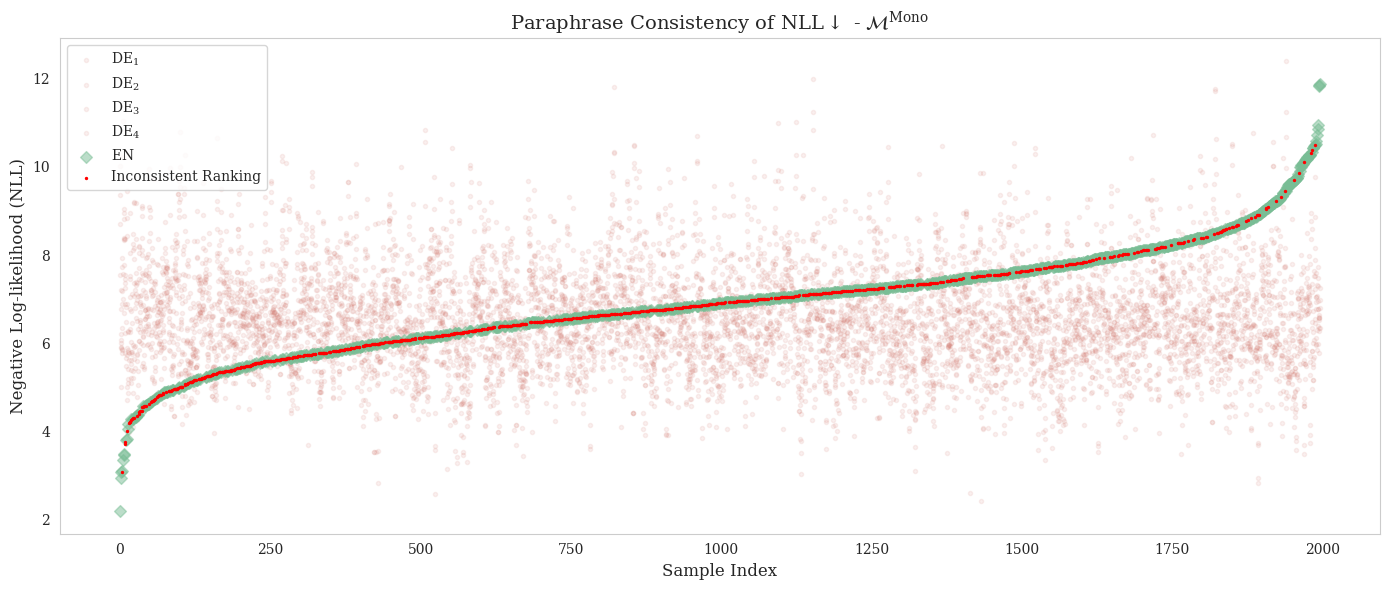}
    \includegraphics[width=\linewidth]{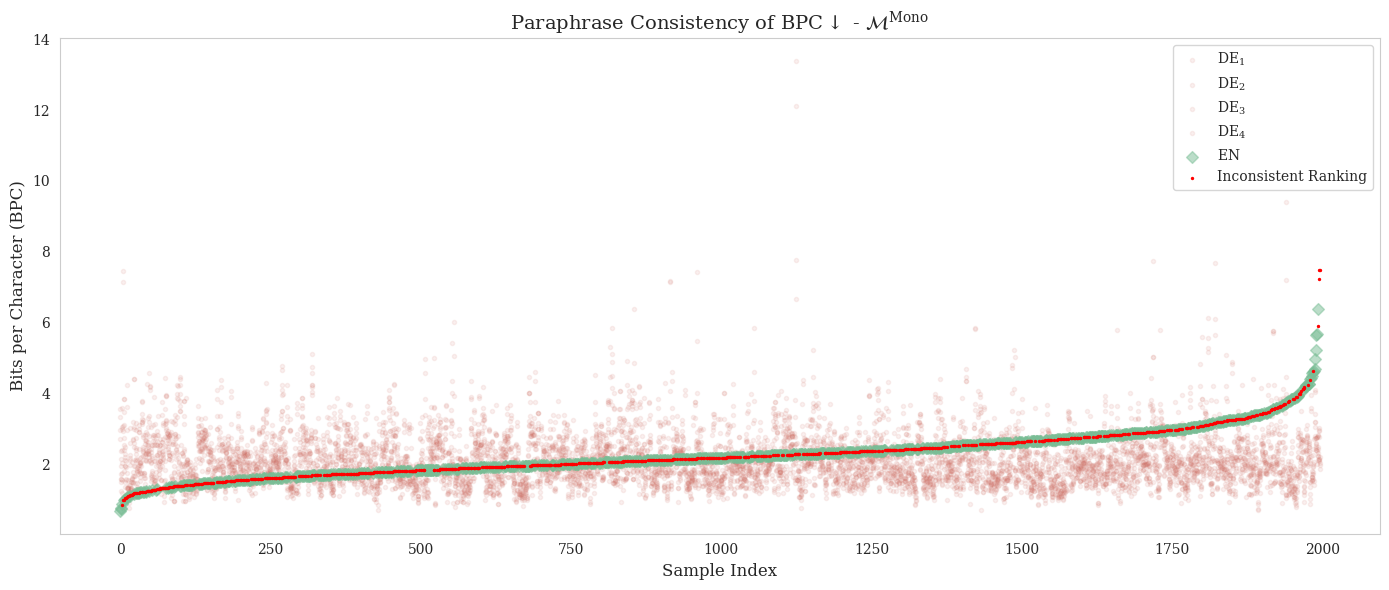}
    \includegraphics[width=\linewidth]{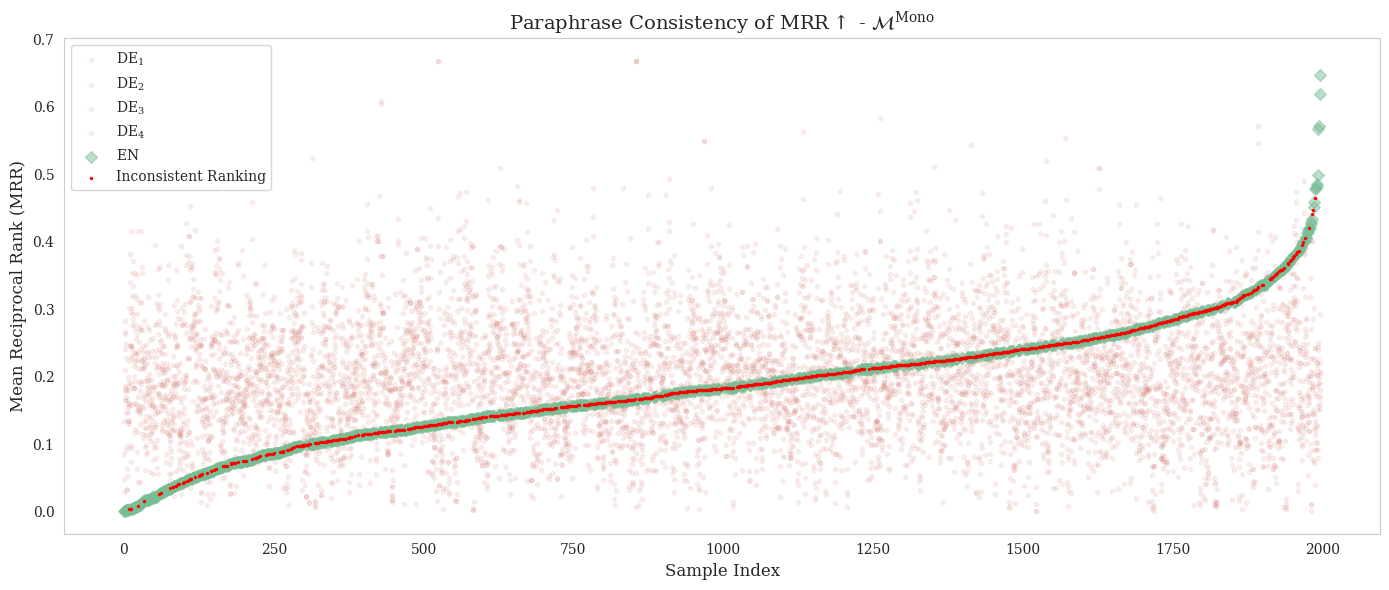}
    \caption{Consistency of NLL, BPC, and MRR using $\mmono_{\text{DE}}$ and $\mmono_{\text{EN}}$ trained on EuroParl.}
    \label{fig:sens-ep-mono}
\end{figure*}

\end{document}